\documentclass{article}
\usepackage{spconf,amsmath,graphicx}
\usepackage{hyperref}

\title{Balanced Face Dataset: Guiding StyleGAN to Generate Labeled Synthetic Face Image Dataset for Underrepresented Group}
\name{Kidist Amde Mekonnen \thanks{The project was completed with the guidance and mentorship of Prof. Kostas Daniilidis, and the author expresses gratitude for the necessary resources and support provided by the professor.}}
\address{AIMS-AMMI, Rwanda\\{
kmekonnen@aimsammi.org \textbar\  kidistamdie@gmail.com}}

\begin{document}
%\ninept
%
\maketitle
\begin{abstract}
For a machine learning model to generalize effectively to unseen data within a particular problem domain, it is well-understood that the data needs to be of sufficient size and representative of real-world scenarios. Nonetheless, real-world datasets frequently have overrepresented and underrepresented groups. One solution to mitigate bias in machine learning is to leverage a diverse and representative dataset. Training a model on a dataset that covers all demographics is crucial to reducing bias in machine learning. However, collecting and labeling large-scale datasets has been challenging, prompting the use of synthetic data generation and active labeling to decrease the costs of manual labeling. The focus of this study was to generate a robust face image dataset using the StyleGAN model. In order to achieve a balanced distribution of the dataset among different demographic groups, a synthetic dataset was created by controlling the generation process of StyleGaN and annotated for different downstream tasks.

The dataset will be released on \href{https://github.com/kidist-amde}{Github}.
\end{abstract}
\section{Introduction}
\label{sec:intro}

Deep learning has proven to be a successful approach in several machine learning domains, such as Computer vision, Natural language processing, and Speech processing.\cite{krizhevsky2017imagenet,vaswani2017attention,radford2019language,graves2005framewise,hinton2012deep}. However, one of the limitations of deep learning is that it requires a lot of data and is often labeled datasets\cite{lecun2015deep,Goodfellow-et-al-2016,shankar2017no,ng2016nuts,hassabis2017neuroscience}. \\
Advancement in high-performance hardware has made it feasible to train large deep-learning models. These models generally consist of a large number of trainable parameters and require a vast dataset for their training. In the domain of facial images, the large-scale datasets currently employed are frequently gathered without considering the demographic distribution, leading to biased data. The training dataset used to train these large models is often lacking in transparency and careful discretion, resulting in a lack of geodiversity that inadvertently produces data resulting in gender, ethnic, and cultural biases. When we disregard how data is collected, processed, and organized, it leads to uneven distribution and bias in the dataset, ultimately causing a biased model when trained on the dataset\cite{wang2021deep,buolamwini2018gender,torralba2011unbiased,gebru2021datasheets,karkkainen2019fairface}. 
%\\
%Deep convolutional neural networks have been successfully used for many computer vision problems. However, training these networks needs higher-resolution images \cite{dodge2016understanding}.

Generative Adversarial
Networks (GANs) based models have been used to generate synthetic datasets\cite{aggarwal2021generative,pan2019recent}. GANs \cite{goodfellow2020generative}  are deep
generative models that have two networks namely a generator, and a discriminator which compete with one against the other. The task of the generator network is to generate synthetic data that imitate the real data to trick the discriminator and the role of the discriminator network is to distinguish the fake image from the real one.
However, GANs are limited to small dataset sizes, with low-resolution
image generation. Due to this generating a high-resolution image has been a challenging task.
Controlling the attributes of the generated image also poses another challenge to image generation.

StyleGAN\cite{karras2019style} is an extension to the GAN architecture that advances the generator model. StyleGAN tackled
the traditional GAN model challenges by building Style based GAN architectures to generate high-quality images and disentangling the latent factor of variation. StyleGAN has also been trained on the Flickr-Faces-HQ dataset and it has shown that it can generate high-quality (1024 x 1024) realistic synthetic human face images. Disentangling the latent factor of variation in StyleGAN has enabled the guided generation of images.

This research aims to investigate how the distribution of datasets across different demographic groups impacts the model bias. Specifically, we seek to answer the following research questions:
\begin{itemize}
\item  How does an uneven distribution of datasets across different demographic groups contribute to model bias?
\item   What are the cost-effective ways to create an evenly distributed dataset across different demographic groups?
\item   Can generating synthetic datasets help reduce the biases of ML models?
\end{itemize}
To attain the research objectives, a novel approach known as "demographic-based dataset balancing" is proposed, which aims to balance the dataset across different demographic groups to ensure equal representation.  

The ultimate goal of this research is to shed light on the significance of a balanced dataset in mitigating model bias and to develop feasible solutions that can be implemented in practical applications.

In this work, the images have been generated by guiding StyleGAN to generate faces where the face
images are uniformly distributed across different demographic groups.

 This paper makes the following important contributions:
\begin{itemize}
 \item Highlighting and demonstrating the uneven distribution of StyleGAN's face image synthesis, which can lead to fairness concerns by generating biased synthetic data.
\item  Proposes a strategy to address the issue of uneven generation of data samples from different groups by guiding styleGAN in image synthesis. 

\item Generation of a high-quality and large-scale face image dataset using StyleGAN. The dataset is evenly distributed across different demographic groups, making it a valuable resource for researchers in the field of computer vision.
\item Annotating the generated dataset for several downstream tasks, such as eye state, smile, and demographic groups. This annotated dataset can be used to improve fairness and accuracy in facial recognition systems, as well as help researchers working on computer vision tasks such as emotion recognition and facial expression analysis.
\end{itemize}
The paper starts with a brief introduction to the work. The subsequent sections are organized as follows: In section two, the study of previous works on image generation, active labeling, and reducing bias in machine learning is presented. Section three outlines the research methodology employed. The experimental results are showcased in section four. Furthermore, sections five and six dedicated discussions on future work and the conclusion of the study, respectively.

\section{Related Work}
\label{sec:related-work}

Obtaining extensive labeled datasets to be used for the training of machine learning models has posed a difficulty. To address this issue, active labeling has been used to automatically label collected images. However, collecting the images themselves remains a challenge. Another approach that has been proposed is to use Generative Adversarial Networks to generate synthetic images, which can then be labeled using active labeling techniques \cite{chen2020active}.

GANs are deep neural networks that can be trained in adversarial settings and have been used to generate realistic images\cite{goodfellow2020generative}. However, it has been shown that in the original implementation of GANs it is hard to control the generation of images. Conditional GANs are introduced to bias the GANs to generate a specific class\cite{mirza2014conditional}, but training these models to generate high-resolution images poses another challenge. StyleGAN\cite{karras2019style} is a generative model, which enables unsupervised separation of high-level attributes of generated images and intuitive control over the synthesis of images.

Often, face image datasets exhibit an overrepresentation of some groups and an underrepresentation of others. Failure to pay adequate attention to data collection, processing, and organization can result in a lack of geodiversity, leading to biased data that perpetuates gender, ethnic, and cultural biases. Ultimately, biased training data is the primary cause of bias in AI systems\cite{buolamwini2018gender,crawford2021excavating,bolukbasi2016man,rashkin2017truth}. When training GANs using these biased datasets, the trained model also will be biased\cite{jain2020imperfect}.
One solution that has been proposed to deal with bias in GANs was controlled manipulation of specific image characteristics\cite{denton2019detecting}.
Here a vector pointing in the direction of the specific attributes is used to guide the generation of the images. This work overlaps with our method in the way of using a vector to guide the generation of images by
GANs. The other solution proposed was auxiliary classifier GAN which strives for demographic parity
or equality of opportunity\cite{sattigeri2019fairness}. 

The proposed solution enables the generation of uniform distribution
of faces across various demographics. The generator model was trained to learn to generate a fairer
dataset in the original input feature space. Another paper \cite{kenfack2021fairness} discusses the fairness concerns of Generative Adversarial Networks (GANs) and shows that GANs models may inherently prefer certain groups during the training process, which can lead to representation bias during the testing phase. The authors propose two solutions to address fairness concerns in generative adversarial networks (GANs). The first solution is using Conditional GANs as a solution to address the fairness concerns of GANs. This involves conditioning the generator and discriminator on the group label in addition to the class label. By doing so, the generator can generate samples that are homogeneously distributed across different groups during the testing phase. However, this solution needs to be explored further as biases may shift to subgroups. The second solution is an ensemble learning approach, where multiple generators are trained and combined to generate data with an equal representation rate of different groups. Their paper focused on  addressing fairness concerns in GANs while this  work focused on dataset creation. 

(Kärkkäinen et al., 2019)\cite{karkkainen2019fairface} addresses the issue of racial bias in existing public face datasets, which are predominantly biased toward Caucasian faces, leading to inconsistent model accuracy and adversely affecting research findings. These authors proposed a solution to mitigate demographic bias in existing public face datasets by constructing a novel face image dataset with a balanced race composition. The dataset contains over 100,000 images with a balanced race composition. Unlike this work, the dataset was created by collecting images from the YFCC-100M Flickr dataset and labeling them with race, gender, and age groups. 

 In their work, Qi and colleagues \cite{mao2019mode} introduced a regularization technique for mode seeking, aimed at mitigating the mode collapse issue in CGANs. Nevertheless, their method does not align with the objective of generating fair data from distinct groups since it does not guarantee diversity in the generated samples. This means that even a varied set of samples may still lack representation from underrepresented groups.

 Xu et al. \cite{xu2018fairgan}
 developed a GAN architecture called FairGAN that generates data that is considered "fair" based on statistical parity. Their method aims to remove any sensitive attribute information from the generated data while preserving as much information as possible. They incorporated an additional discriminator that differentiates whether the generated samples are from the protected or unprotected group. Although their work also tackles fairness in GANs, their objective is different  than this work.
\section{Method}
\label{sec:method}
The process of generating and labeling synthetic face images is divided into different pipelines. The pipelines are executed sequentially. This section provides a detailed explanation of each pipeline, and the result of each pipeline is presented in the following section.

\subsection{Guided Image Generation and Labeling}
The objective of this project is to generate images that are uniformly distributed across different demographic groups. To achieve this, the initial step involved training a model to classify images into different demographic groups. The model was trained to identify the following demographic groups: Asian, Black, Indian, White, and Others. The setup is robust and can be extended to any number of groups.

To train a race classifier, we utilized the UTKFace \cite{zhifei2017cvpr} image dataset, which comes equipped with a race label. The VGGFace \cite{Parkhi2015DeepFR} model was fine-tuned to create the race classifiers.  The RaceClassifier model was then used to identify the race of images during the image generation process using StyleGAN. This model is referred to as the RaceClassifier throughout this paper.

\subsubsection{Training classifier from face images}
The deep neural network model takes an image as input and predicts the corresponding race label for that particular image.

\begin{figure}[htb]

\begin{minipage}[b]{1.0\linewidth}
  \centering
  \centerline{\includegraphics[width=8.5cm]{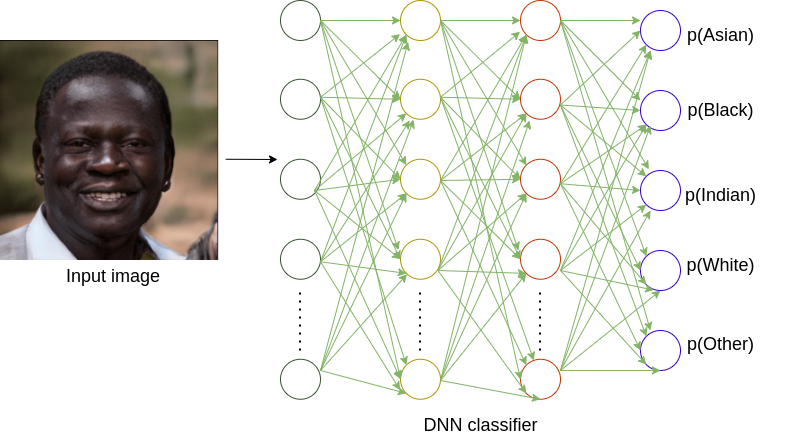}}

\end{minipage}
\caption{Race classifier model.}
\label{fig:res}
\end{figure}

Let X be a set of image features in the UTKFace image dataset and Y be a set of labels associated with the image features. The race classifier  model is a function that accepts input images $x_i$ $\in X$ and predicts a target $y_i$ $\in Y$. Mathematically the model corresponds to the following equation. 
\begin{equation}
RaceClassifier(x_i) = P_{\theta}(y_i| x_i).\ \ x_i \in X \ and \  y_i \in Y
\end{equation}
The pre-trained StyleGAN face image generator model was utilized to produce $180,000$ face images.\\
Let $Z$ represent the latent input space of the StyleGAN model. $Z \in R^{512}$ and normally distributed around zero with standard deviation of 1.
\begin{equation}
Z\in R^{512},\ \ \  z_i \sim \mathcal{N}(0, 1) 
\end{equation}
To generate each image, $Z_i$  fed to StyleGAN. 
\begin{equation}
\hat{x}_i = StyleGAN(z_i)
\end{equation}
\subsubsection{ Generating and labeling high-resolution face image }

\begin{figure}[htb]
\begin{minipage}[b]{1.0\linewidth}
  \centering
  \centerline{\includegraphics[width=8.5cm]{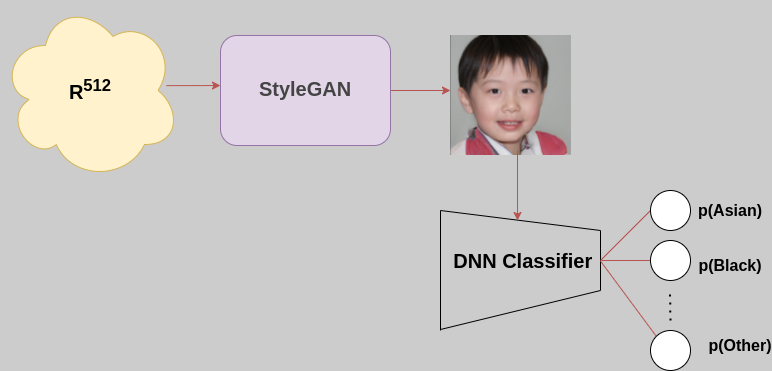}}
\end{minipage}
\caption{{Face image generation and race classification.}}
\label{fig:gen}
\end{figure}

Figure \ref{fig:gen}  illustrates a process in which random vectors (Z) are initially generated from a normal distribution of $R^{512}$ dimensions, and then used as input to StyleGAN to generate face images. These images are subsequently fed into the RaceClassifier model to obtain their corresponding race label. The generated image, input vector, and race label are then mapped and stored for future use. In order to maintain a uniform distribution of images across all race groups, only 10,000 images per race were saved.\\
It is assumed that the output generated by the StyleGAN model will exist in the same vector space as the input for the RaceClassifier model. To ensure this, the resolution of the images was reduced before being fed into the RaceClassifier model.

 \subsection{Training Logistic regression that classifies StyleGAN input space}
Manipulation of image generation is possible using the StyleGAN model. Consequently, The factors  of variation are disentangled, which means moving a small distance in a certain direction changes only a specific or small set of features. This fact was proved in this work while generating images for some demographic groups. 
Using the dataset generated ($R^{512}$ input vector, image, race label), for each demographic group, one-vs-rest logistic regression model was trained to classify the input space of the StyleGAN model into positive and negative regions. This model is referred to as the LogisticClassifier. 

\begin{figure}[htb]
\begin{minipage}[b]{1.0\linewidth}
  \centering
  \centerline{\includegraphics[width=8.5cm]{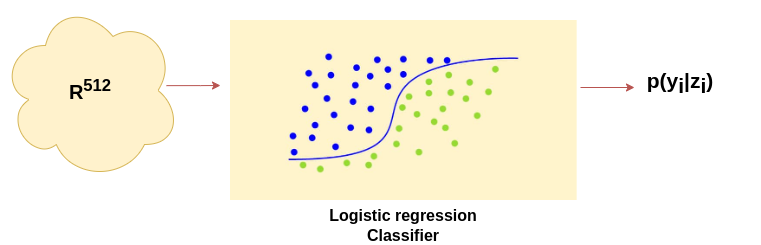}}
\end{minipage}
 \caption{StyleGAN input space classifier. Multiclass Classification One vs All (One vs Rest)}
\label{fig:logstic}
\end{figure}

\subsection{The parameter vector and hyperplane}
\label{heading:sec4}
For each demographic group, the LogisticClassfier’s parameters vector is used to compute a vector perpendicular
to the hyperplane of the classification. This vector is used to guide the generation of the images
from StyleGAN by maximizing the probability of generating a face image with a given demographic group. The RaceClassifier model is used to verify if the generated images are from the group of interest. If the
images are not from the group they will be discarded, but if they are from the group of interest the
triple (input, image, race) will be saved for later use.

\begin{figure}[htb]
\begin{minipage}[b]{1.0\linewidth}
  \centering
  \centerline{\includegraphics[width=8.5cm]{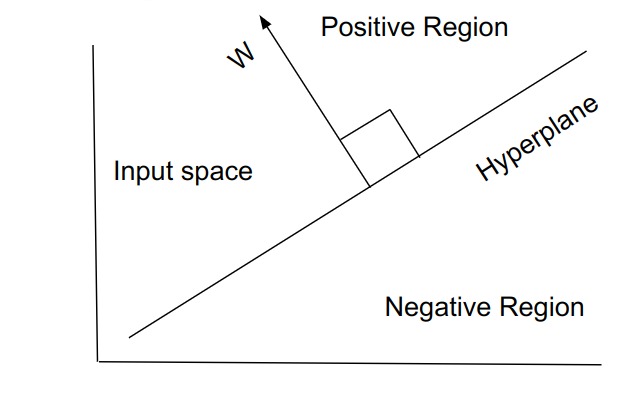}}
\end{minipage}
\caption{Parameter vector pointing to the positive direction of race attribute.}
\label{fig:latent}
\end{figure}

This section discusses how to generate darker skin face images by guiding the styleGAN model. The results can be extended to control many features of generated face images. 

Let $ Z\subset R ^{512}$ be our input space of the StyleGAN and let 

$f(z)$ represents the parameterized LogisticClassifier of the darker skin face. 
\begin{equation}
f_\theta(z_i) = \sigma(<\theta, z_i>)
\end{equation}
We are looking for a directional vector that maximizes the probability of finding the latent feature that can generate a darker skin face image when fed to the StyleGAN model. The function $f(z)$ has been trained and can map any point in $Z $ to $(0, 1)$.

On the classification hyperplane, we have the following equation. 
\begin{equation}
	\begin{aligned}	
f_\theta(z_i) = \sigma(<\theta, z_i>)=0.5\\
\implies <\theta, z_i> = 0
\end{aligned}
\end{equation}
Since we want to maximize the probability of finding the latent feature vector which maximizes $f(z)$, we can write the objective as follows. 
\begin{equation}
\operatorname*{argmax}_{z_i}  f_\theta(z_i)
\end{equation}
To find a unique solution we have restricted our choice of $z_i$ to have an $L2 $ norm of $1$. With this constraint, we can write the problem as follows. 

\begin{equation}
	\begin{aligned}
	\operatorname*{argmax}_{z_i}  f_\theta(z_i)&= \operatorname*{argmax}_{z_i}  ||\theta||_2||z_i||_2cos(\alpha)\\
	z_i^*&=\frac{\theta}{||\theta||_2}
	\end{aligned}
\end{equation}
Thus the direction pointed by the parameters vector is perpendicular to the hyperplane \ref{fig:latent}  and it is also the direction that maximizes finding the latent feature vectors that will generate darker skin face images.

\subsection{Training models for downstream tasks } 
Deep neural networks are utilized for performing various downstream tasks, including emotion classification, eye state detection, smile detection, mask detection, and gender recognition. CNN models have been developed for this purpose, and the CelebA \cite{liu2015faceattributes} and \href{https://www.kaggle.com/c/challenges-in-representation-learning-facial-expression-recognition-challenge/data}{Kaggle FER2013} datasets were employed for training the models. These trained models are referred to as DownstreamModels in this study.
\subsection{Labeling the generated images}
The final pipeline of this work involves using DownstreamModels to label the dataset. It has been observed that the models occasionally make incorrect predictions, particularly when the predicted class has a low probability. Therefore, images that were predicted with lower probabilities were selected and manually labeled.

\section{Experimental Results}
\subsection{RaceClassifier}
To construct a RaceClassifier model, a pretrained VGGface model was used. The UTK-Face dataset was used in the training process with an Adam optimizer and a learning rate of 0.001. The model was trained for 50 epochs with early stopping to prevent overfitting that occurs after a few epochs. The model's performance on the training, validation, and test sets is presented in the table below.
\begin{table}[h]
\begin{center}
 \begin{tabular}{|c| c |c |c|} 
 \hline
  & Training & Validation & Test \\ [0.5ex] 
 \hline
 Accuracy & 79.93\% & 80.86\% & 80.12\% \\ 
 \hline
 Loss & 0.5545 & 0.5466 & 0.5753 \\
  \hline
\end{tabular}
\end{center}
  \caption{Performance of RaceClassfier model\label{tab:table1}}
\end{table}

\subsection{StyleGAN image Generation}
The following images are generated randomly by the StyleGAN model. First a random vector from the $R^{512}$ space is generated. All components of the random vector are drawn from standard normal distribution.

\begin{figure}[htb]
\begin{minipage}[b]{1.0\linewidth}
  \centering
  \centerline{\includegraphics[width=8.5cm]{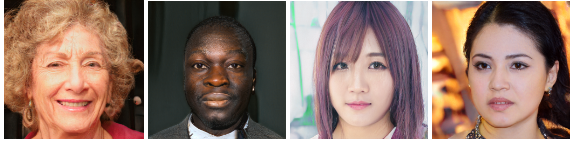}}
\end{minipage}
\caption{Sample images generated by StyleGAN model.}
\label{fig:img}
\end{figure}

When there is no guiding, the distribution of images being generated is biased towards Caucasian face images. To make concrete justification we have randomly generated $10,000$ face images using StyleGAN. Table \ref{tab:table2} shows the number of face images for each demographic groups.
\begin{table}[h]
\begin{center}
 \begin{tabular}{|c| c| c| c| c |c|} 
 \hline
 Class & Asian & Black & Indian & White & Others \\ [0.5ex] 
 \hline
 No. of images  & 882 & 449 & 975 & 6837 & 867 \\ 
  \hline
 Percentage \%   &8.82\%  & 4.49\% &   9.75\%&  68.37 \%& 8.67\% \\ 
  \hline
\end{tabular}
\end{center}
  \caption{Generated images distribution across different demographic groups. Most of the time the model generates Caucasian face images.\label{tab:table2}}
\end{table}
\subsection{Guided StyleGAN image Generation}
As it has been discussed in Section \ref{heading:sec4}  the direction pointed by the parameter vector is perpendicular to the hyperplane and it is also the direction that maximizes finding the latent feature vectors that will generate the target face images. Using this fact, a random vector was randomly chosen from the latent input space, and the parameter vector of the LogisticClassifier add to the random vector before feeding it to StyleGAN. 

Table \ref{tab:table3} shows the percentage of the images for each demographic group when guiding is used. 1000  images per demographic have been generated. As it has been shown in the table, the percentage of the images for each underrepresented group has increased. 
\begin{table}[h]
\begin{center}
 \begin{tabular}{|c |c |c |c|} 
 \hline
 Class & Asian & Black & Indian\\ [0.5ex] 
 \hline
 No. of images  & 423 & 633 & 259  \\ 
 \hline
 Percentage   & 42.3\% & 63.3\% & 25.9\%  \\ 
  \hline
\end{tabular}
\end{center}
\caption{Distribution of generated images when guiding is applied. The bias was highly reduced. }.\label{tab:table3}
\end{table}

\begin{figure}[htb]
\begin{minipage}[b]{1.0\linewidth}
  \centering
  \centerline{\includegraphics[width=8.5cm]{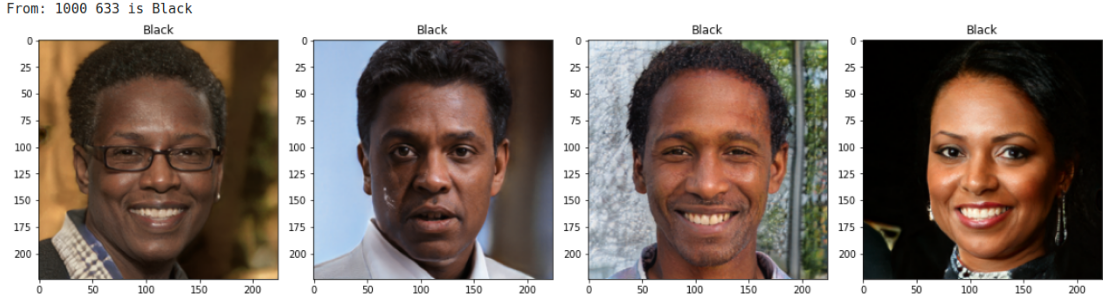}}
    \centerline{\includegraphics[width=8.5cm]{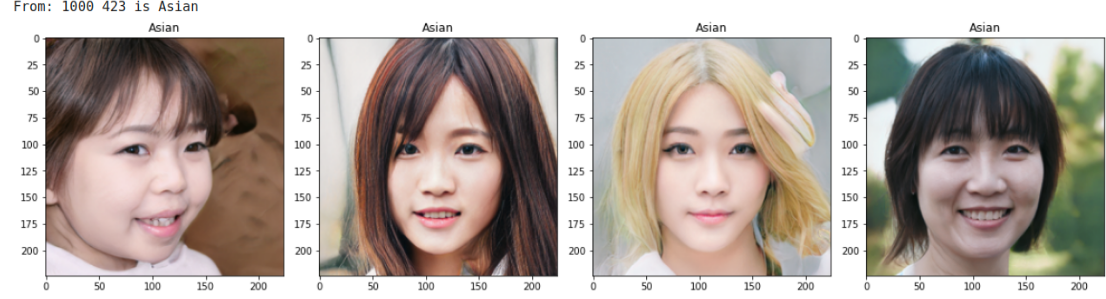}}

\end{minipage}
\caption{ Sample images when guiding StyleGAN model to generate specific demographic group face image.}
\label{fig:face}
\end{figure}

\section{Discussion and Future work}
A large-scale dataset was generated with potential for future improvements. The ability to guide StyleGAN to generate face images with specific skin color was demonstrated, and this approach can be extended to include other facial features, such as happy faces, closed eyes faces, talking faces, engaged faces, etc. Additionally, generating faces with specific features can be accelerated by fixing the latent feature vector around a region with a small radius. This involves randomly sampling a latent vector, adding directional parameters, and fixing the vector around this new point, after which many faces with similar features can be generated by moving a small distance from this new point.

The dataset generated can serve as a baseline for training models and measuring their performance on uniformly distributed datasets for face-related tasks.

\section{Conclusion}
This research presents a large-scale labeled dataset that is evenly distributed among different demographic groups. The process of generating the synthetic face images dataset and labeling the dataset is also described. 
 Training the classifier on this expanded and diverse collection of dataset helps minimize biases against underrepresented groups. Consequently, this work represents a step toward addressing algorithmic bias.
\section{ETHICS STATEMENT}

This research focuses on mitigating bias in machine learning,  particularly arising from the utilization of unbalanced datasets in training.   The approach involves guiding the image generation process of a StyleGAN model to produce a diverse collection of facial images. This collection ensures equal representation across different demographic groups. Throughout our study, we are dedicated to identifying and rectifying any unintended biases that might emerge during the phases of dataset generation and annotation.   Additionally, the potential for biases within the model's architecture and training procedures is acknowledged. Recognizing these possibilities is crucial before engaging in model development, deployment, or enhancement.

The research incorporates these concerns into the algorithms, aimed at enhancing the balanced generation of StyleGANs across diverse demographic categories. We acknowledge the limitations inherent in real-world datasets and underline the significance of training on a comprehensive dataset encompassing all demographics to foster fairness and inclusivity. Leveraging synthetic data generation assists in achieving equitable distributions and minimizing the need for labor-intensive manual labeling. Ethical considerations steer us toward ensuring privacy safeguards and preventing the propagation of harmful depictions. 

The ultimate goal is to contribute to responsible AI advancement, prioritizing societal benefits.

\section{Acknowledgment}
This paper was initially completed by the author as a requirement for obtaining an MSc in Machine Intelligence degree at AMMI-AIMS Rwanda, with a submission date of 30 Mar 2021. An earlier version of the paper was presented at the \href{https://appliedmldays.org/events/amld-africa-2021/speakers/kidist-amde-mekonnen}{AMLD Africa 2021} on September 3rd on the AI for Social Good track. The project was completed with the guidance and mentorship of Prof. Kostas Daniilidis, and the author expresses gratitude for the necessary resources and support provided by the professor.

\bibliographystyle{IEEEbib}
\bibliography{strings,main}

\end{document}